\def\year{2022}\relax
\documentclass[letterpaper]{article} 
\usepackage{aaai22}  
\usepackage{times}  
\usepackage{helvet}  
\usepackage{courier}  
\usepackage[hyphens]{url}  
\usepackage{graphicx} 
\usepackage{natbib}  
\usepackage{caption} 
\DeclareCaptionStyle{ruled}{labelfont=normalfont,labelsep=colon,strut=off} 
\usepackage{algorithm}
\usepackage{algorithmic}
\usepackage{amssymb}
\usepackage{xcolor}

\usepackage{newfloat}
\usepackage{listings}
\floatstyle{ruled}
\newfloat{listing}{tb}{lst}{}
\floatname{listing}{Listing}

\title{A Learning Based Framework for Handling Uncertain Lead Times in Multi-Product Inventory Management}
\author {Hardik Meisheri~\textsuperscript{\rm 1}, Somjit Nath~\textsuperscript{\rm 1}, Mayank Baranwal~\textsuperscript{\rm 1, \rm 2}, Harshad Khadilkar~\textsuperscript{\rm 1, \rm 2}}

\affiliations{
\textsuperscript{\rm 1}TCS Research, Mumbai\\
\{hardik.meisheri, somjit.nath, baranwal.mayank, harshad.khadilkar\}@tcs.com\\
\textsuperscript{\rm 2}IIT Bombay, Mumbai \\
\{mbaranwal, harshadk\}@iitb.ac.in

}

\begin{document}

\maketitle

\begin{abstract}
Most existing literature on supply chain and inventory management consider stochastic demand processes with zero or constant lead times. While it is true that in certain niche scenarios, uncertainty in lead times can be ignored, most real-world scenarios exhibit stochasticity in lead times. These random fluctuations can be caused due to uncertainty in arrival of raw materials at the manufacturer's end, delay in transportation, an unforeseen surge in demands, and switching to a different vendor, to name a few. Stochasticity in lead times is known to severely degrade the performance in an inventory management system, and it is only fair to abridge this gap in supply chain system through a principled approach. Motivated by the recently introduced delay-resolved deep Q-learning (DRDQN) algorithm, this paper develops a reinforcement learning based paradigm for handling uncertainty in lead times (\emph{action delay}). Through empirical evaluations, it is further shown that the inventory management with uncertain lead times is not only equivalent to that of delay in information sharing across multiple echelons (\emph{observation delay}), a model trained to handle one kind of delay is capable to handle delays of another kind without requiring to be retrained. Finally, we apply the delay-resolved framework to scenarios comprising of multiple products subjected to stochasticity in lead times, and elucidate how the delay-resolved framework negates the effect of any delay to achieve near-optimal performance.
\end{abstract}

\section{Introduction \& Related Work}
Inventory replenishment~\cite{axsater2015inventory} is one of the key factors determining efficient flow of goods through the entire supply chain system. Inventory replenishment primarily concerns with designing strategies that govern movement of inventory from reserve storage (e.g. a warehouse) to primary storage (e.g. a retail outlet). Devising better strategies for replenishment translates directly to increased profit margins for a business. Thus, it is only fair that the problem of inventory replenishment remains one of the most studied problems in the operations research community.

There are several factors that impact the efficacy of inventory replenishment - stochastic demands~\cite{lewis2012demand}, limited capacity of vehicles~\cite{sindhuchao2005integrated}, finite shelf life of goods~\cite{haijema2013new}, cross-product constraints~\cite{minner2010periodic}, limited holding capacities at primary and secondary storage~\cite{bertsimas2016inventory}, uncertainty in manufacturing or procurement lead times~\cite{dolgui2013state}, and poor end-to-end visibility~\cite{goh2009supply}, i.e., unavailability of precise, real-time info regarding current stock levels to suppliers. The problem becomes particularly critical when the replenishment strategies need to be designed concurrently for a very large number of products.

Existing supply-chain management strategies incorporate one or only a few factors into their decision making, resulting in an inefficient and sub-optimal replenishment system. For instance, the economic-order-quantity model and the dynamic-economic-lotsize model, two of the most commonly used replenishment policies work under the assumption of deterministic demands~\cite{zipkin2000foundations}. Consequently, a robust linear programming framework was developed~\cite{bertsimas2004robust} to address the shortcomings of the aforementioned policies. The framework, originally designed to handle single-stage single-product, is computationally scalable but cannot handle cross-product constraints and backlogging of excess demands. Subsequently, a more recent series of work  concerns with multi-product, multi-echelon replenishment strategies~\cite{aharon2009robust, akbari2015new, cardenas2014optimal, harifi2020optimization, mousavi2014multi, yang2017integrated}, albeit under assumptions of deterministic or negligible (zero) lead times.

A significant disadvantage with the aforementioned strategies is their dependency on solving robust optimization problems at each time point, which scales poorly with the number of products. In reality, a moderately large retail business may have each of its store selling up to 100,000 product types~\footnote{\url{https://www.retailtouchpoints.com/resources/how-many-products-does-amazon-carry}}. Fueled by the recent advances and successes of modern deep learning algorithms, there is a significant shift towards adopting reinforcement learning (RL) for handling large product size~\cite{Meisheri2021ScalableMI,yan2021reinforcement}. Recall that the ultimate goal for designing better replenishment strategies is to maximize the long-term (discounted) reward subjected to constraints and uncertainties. This makes RL a suitable candidate to help discover optimal strategies in highly dynamic environments. Despite substantial efforts towards increasing the scope of modern inventory management strategies, uncertainty in lead times and poor end-to-end visibility, which are known to result in bullwhip effect~\cite{lee1997bullwhip}, are often ignored largely due to increased complexity of the resulting supply chain system.

This paper fills the key gap in designing replenishment strategies in realistic scenarios comprising of all levels of complexities, i.e., stochastic demands and lead times, cross-product constraints, large number of products, limited shelf-lives and capacities of products, and nonlinear business rewards. The paper adopts RL as its core solution methodology to address scalability and fragility of supply chain system. Interestingly, lead times in delivery and manufacturing can be viewed as \emph{action delays}, where the delay is between the events when the order is placed till it gets procured. Consequently, we equate the effect of uncertain lead times in inventory replenishment problem to learning in delayed environments. The authors in \cite{nath2021revisiting} recently proposed a delay-resolved framework to handle stochastic delays in highly dynamic environments. We leverage a similar framework to augment our RL-based replenishment strategy to incorporate uncertainty in lead times. As a happy consequence, our framework is also capable to handle uncertainty in real-time information sharing across multiple echelons in a supply chain system. The proposed framework advances the current state-of-the-art methodologies in replenishment system by considering multiple realistic scenarios concurrently in an efficient manner.

While the work adopts RL-based solution as its core methodology, yet it promises a significant departure from the existing literature on using RL for optimizing inventory. Other than its ability to handle stochastic lead times and poor end-to-end visibility, the proposed framework is data-efficient on three accounts, none of which has been addressed in the literature~\cite{Meisheri2021ScalableMI}: (a) {\bf No forecasts}: Our framework does not explicitly require the demand forecasts at all times, except for current timestep. Instead, the framework learns to manage replenishment through step-rewards. (b) {\bf No roll-outs}: Since the framework does not require forecasted demands, policy roll-out based on forecasts is impossible. Despite the lack of policy roll-out (noisy information), our framework is capable of generating better strategies for replenishment. (c) {\bf Single agent for different lead times}: Since the framework is based on augmenting past actions to its \emph{information state}, it can handle any finite-amount of delay, and thus a \emph{single} agent can be used to optimize replenishment of a product regardless of its current (stochastic) lead time delay.

\textbf{Deployment in real world: } The problem statement is motivated from real-world scenarios and constraints in a typical supply chain management system. We believe that modeling lead time in a structured manner while taking care of computational requirements can greatly enhance the practical usability of such algorithms. Real-time inference time has been proven to be a bottleneck when dealing with millions of products.

\section{Methodology}
Formulating the Supply Chain problem as a reinforcement learning has been already explored before~\cite{Meisheri2021ScalableMI}. Generally, while modelling delays it has been addressed in the form of a multi agent Reinforcement Learning problem. While that can work well for constant and small number of unique delays, it is not generalizable. One critical failure would be in the case, where there are multiple vendors with stochastic lead times, this model would not be scalbale and robust.

Delay Resolved Algorithms \cite{nath2021revisiting} has been a success in addressing both constant and stochastic delays and can be applied to this scenario directly with few modifications. Delay Resolved Algorithms do not approach this problem by assigning a separate model for every delay, but by modifying the Markov Decision Process (MDP). Generally, in the presence of delays, reinforcement learning does not work as well, as the states are no longer Markov and are affected by actions which are yet to be implemented on the environment. This makes the MDP partially observable and consequently it is difficult for RL algorithms to learn. Delay Resolved Algorithms address this problem by appending the states with an action buffer of the un-implemented actions. The solution, albeit simple, has a few interesting properties. It can be easily used with any RL algorithm and it is theoretically sound, i.e. optimizing for the Mean Squared TD Error in the new augmented MDP leads to the same optimal policy as the original MDP. This can also be extended to stochastic delays by using a constant length for the action buffer such that the delay values are upper bounded by the maximum length of the buffer. Also, Delay Resolved Algorithms have the added advantage of robustness to the size of this buffer as it uses zero padding for the action buffers which does not alter the final outputs of the RL agent. For this paper, we used the delay resolved version of DQN \cite{mnih2015human} to address the problem of action delay which can be interpreted to be the lead time for the products that is inherently present in our environment. 

The supply chain replenishment can be formulated into MDPs with continuous state space and discrete actions as described in \cite{Meisheri2021ScalableMI}. Decisions for each product are taken independently while supplying global information about constraints in the states and rewards. State-space is presented in table~\ref{tab:ewrl}. $x_i(t)$ denotes the current inventory level in the store for $i^{th}$ product at time step $t$. Metadata for each product is encapsulated by features $v_i$, $c_i$ and $T_i$ which denotes, volume, weights and shelf life of $i^{th}$ product. As mentioned before we only require a forecast for the next time step that is denoted by $\hat{W}_i(t)$. Features ${\bf{v}}^\intercal\,{\hat{\bf{W}}}(t)$ and ${\bf{c}}^\intercal\,{\hat{\bf{W}}}(t)$ provide information about global constraints which are across products and hence help in driving policy towards optimal decisions across products. We also adopt a similar reward structure as mentioned in~\cite{Meisheri2021ScalableMI}. Action space is 14 discrete actions denoting the amount of quantity to be replenished. 

\begin{table}[h]
	\caption{State space representation}
	\label{tab:ewrl}
	\begin{center}
		\begin{tabular}{|c|l|}
			\hline
			Notation & Explanation \\
			\hline
			$x_i(t)$ & Current inventory level \\
			$\hat{W}_i(t)$ & Forecast aggregate orders in $[t,t+1)$ \\
			$v_i$ & Unit volume \\
			$c_i$ & Unit weight \\
			$T_i$ & Shelf-life \\
			${\bf{v}}^\intercal\,{\hat{\bf{W}}}(t)$ & Total volume of forecast for all products \\
			${\bf{c}}^\intercal\,{\hat{\bf{W}}}(t)$ & Total weight of forecast for all products \\
			\hline
		\end{tabular}
	\end{center}
\end{table}

\begin{figure}
    \centering
    \includegraphics[width=0.47\textwidth]{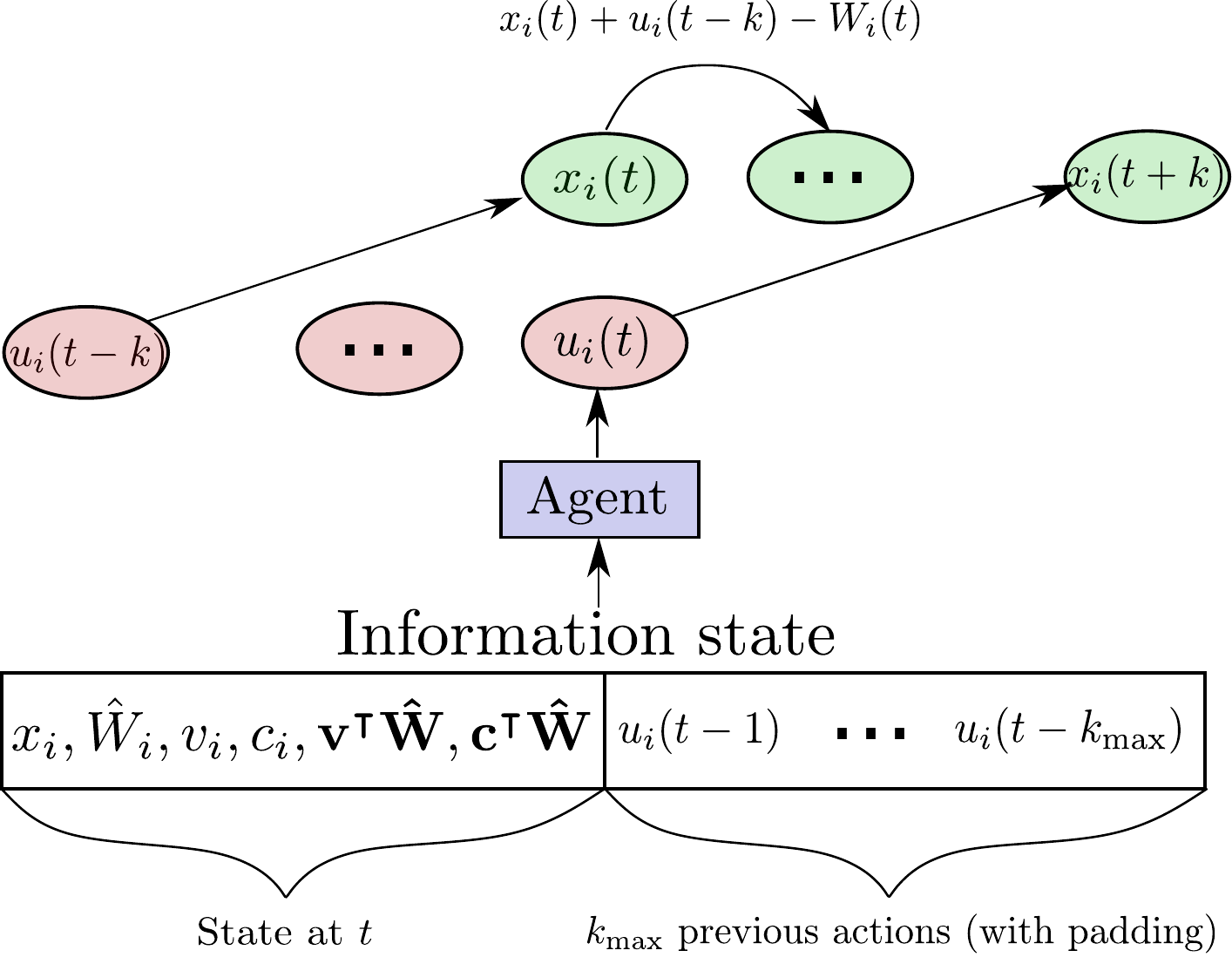}
    \caption{Illustration of the replenishment system with lead times and the information state based RL inputs.}
    \label{fig:system}
\end{figure}

DRDQN in Supply Chains is an extension of the DQN algorithm with augmented states. However, since we have a parallel forward pass for each of the products, the action buffer would also include the delayed actions. Actions for each of the products may be different after applying global constraints such as truck volume and weight capacity. We augment modified actions after applications of global constraints for DRDQN in the state space, whereas original decisions of agents are kept as actions in memory buffer. Information state for DRDQN is shown in Figure~\ref{fig:system}. For stochastic delay cases, we assume that the delay changes only after an episode has been completed. At start of each episode, we sample lead time (delay) uniformly from $(1, k_{max})$. This corresponds to the practical scenario, where at the start of a new season, one has to generate a contract with a vendor with known lead times which remains fixed for a given duration.

In addition, delay-resolved algorithms can be used for both action and observation delays and have equivalent performance for both of them as highlighted in~\cite{nath2021revisiting}. Hence, in the supply chain scenario, Delay Resolved DQN is used to address both lead time delays as well delay in information sharing.

{The proposed framework saves considerable compute time as we do not need to train a model for every lead time. It must be noted that the models trained to account for a specific lead time do not generalize to other lead times. On the other hand, the DRDQN-based framework can be used to account for stochastic delays making it suitable to adapt to different lead times even during the training phase. This makes the proposed framework quite appealing for handling uncertain lead times. Additionally, the complexity of the framework grows linearly with the lead time, resulting in significantly less computational budget as opposed to with the multi-agent RL framework.}

\section{Results and Discussion}

We have used two separate benchmark datasets each with having different characteristics in demand distribution and product metadata with 100 and 220 products respectively~\cite{Meisheri2021ScalableMI} with authors' consent. In our experiments, we consider lead time equal to delays in action implementations. We have used epsilon greedy as exploration strategies and ran each experiment for 10 random seeds for statistical significance. We gauge the performance of our models on the business reward.

\begin{figure}[t]
  \centering
  \includegraphics[width=0.99\linewidth]{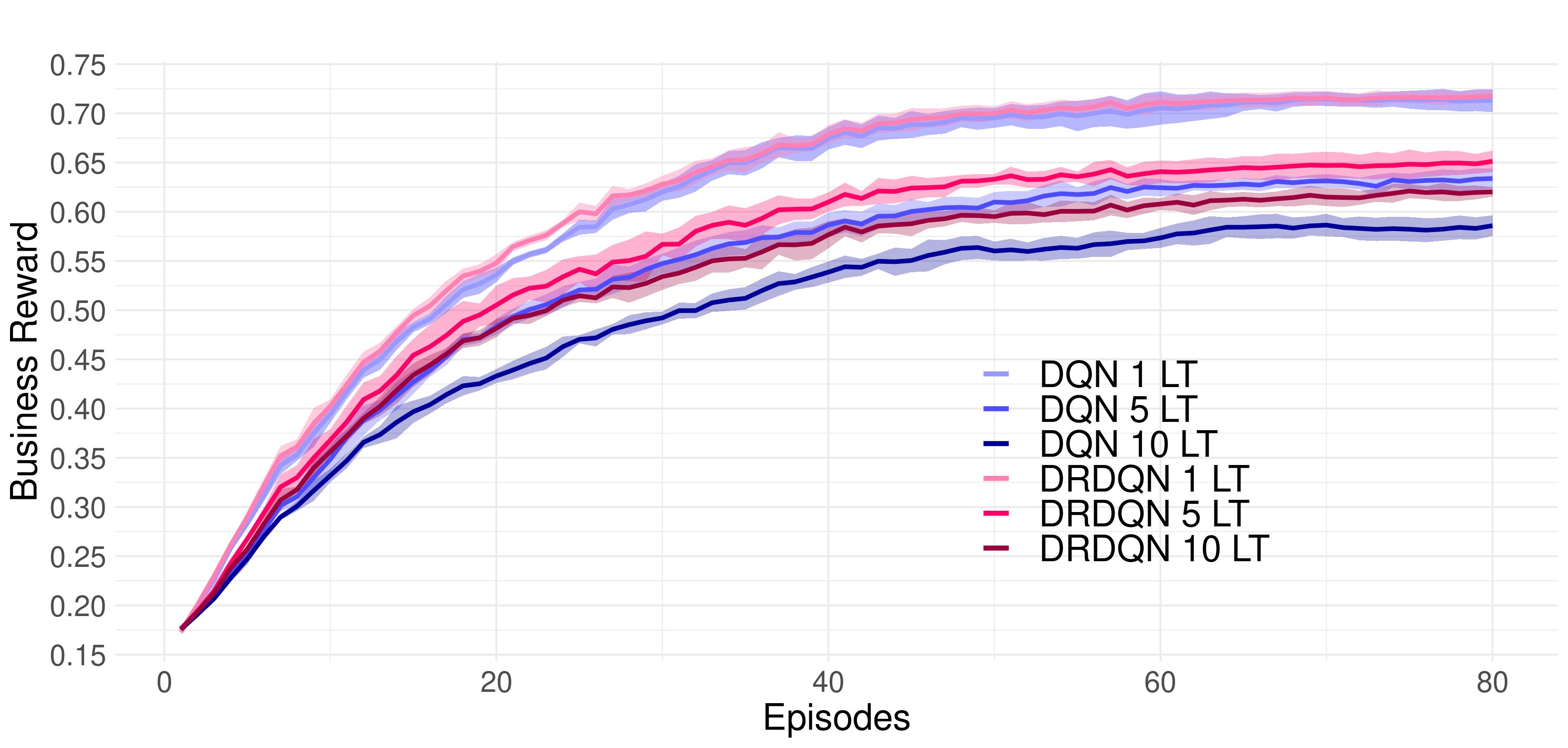}
\caption{Training results over 220 product datasets, solid line represents the mean over 10 random seeds and shaded region denotes 95 percentile confidence interval.}
\label{fig:220_Training}
\end{figure}

\begin{figure}[t]
  \centering
  \includegraphics[width=0.99\linewidth]{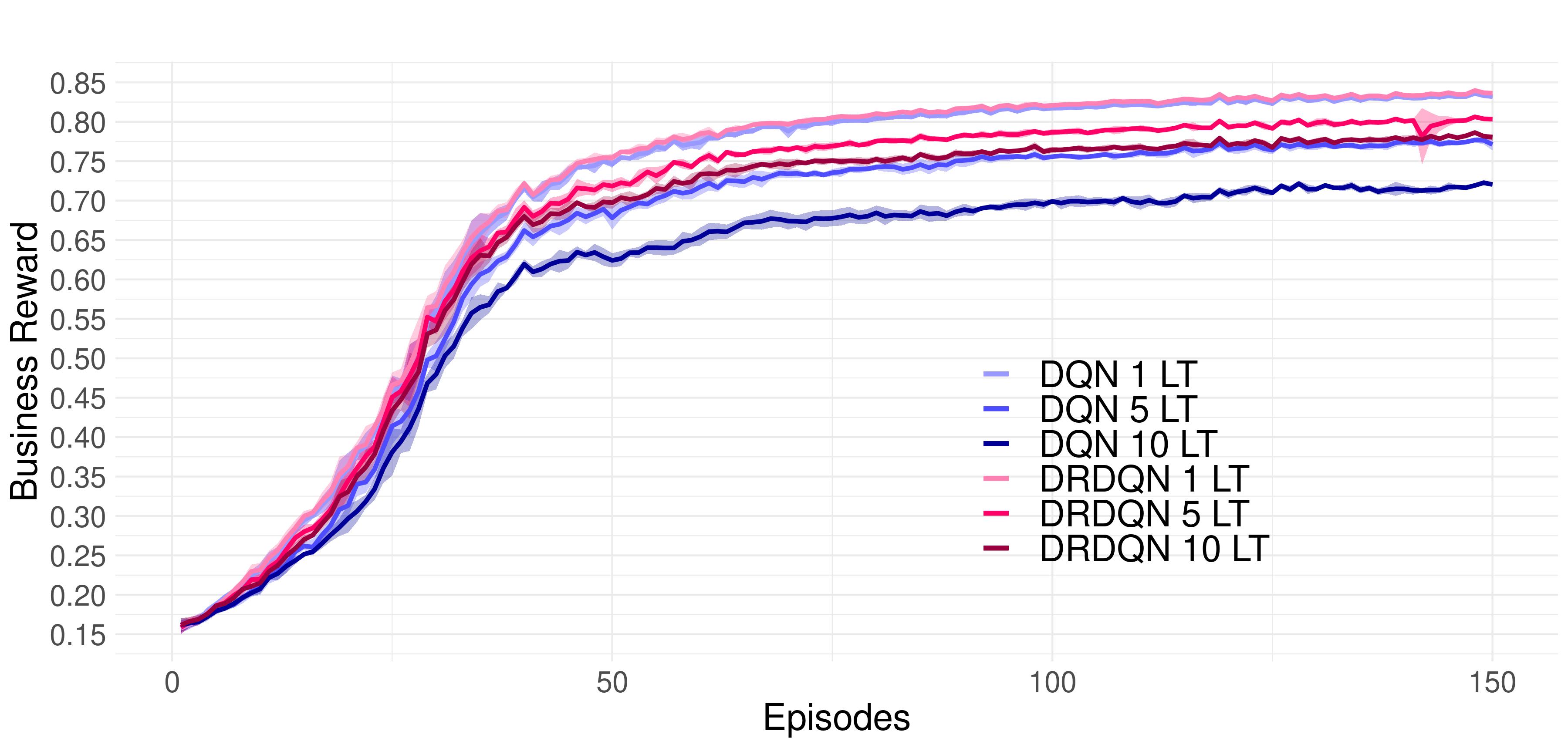}
\caption{Training results over 100 product datasets, solid line represents the mean over 10 random seeds and shaded region denotes 95 percentile confidence interval. }
\label{fig:100_Training}
\end{figure}

\begin{figure}[H]
  \centering
  \includegraphics[width=0.99\linewidth]{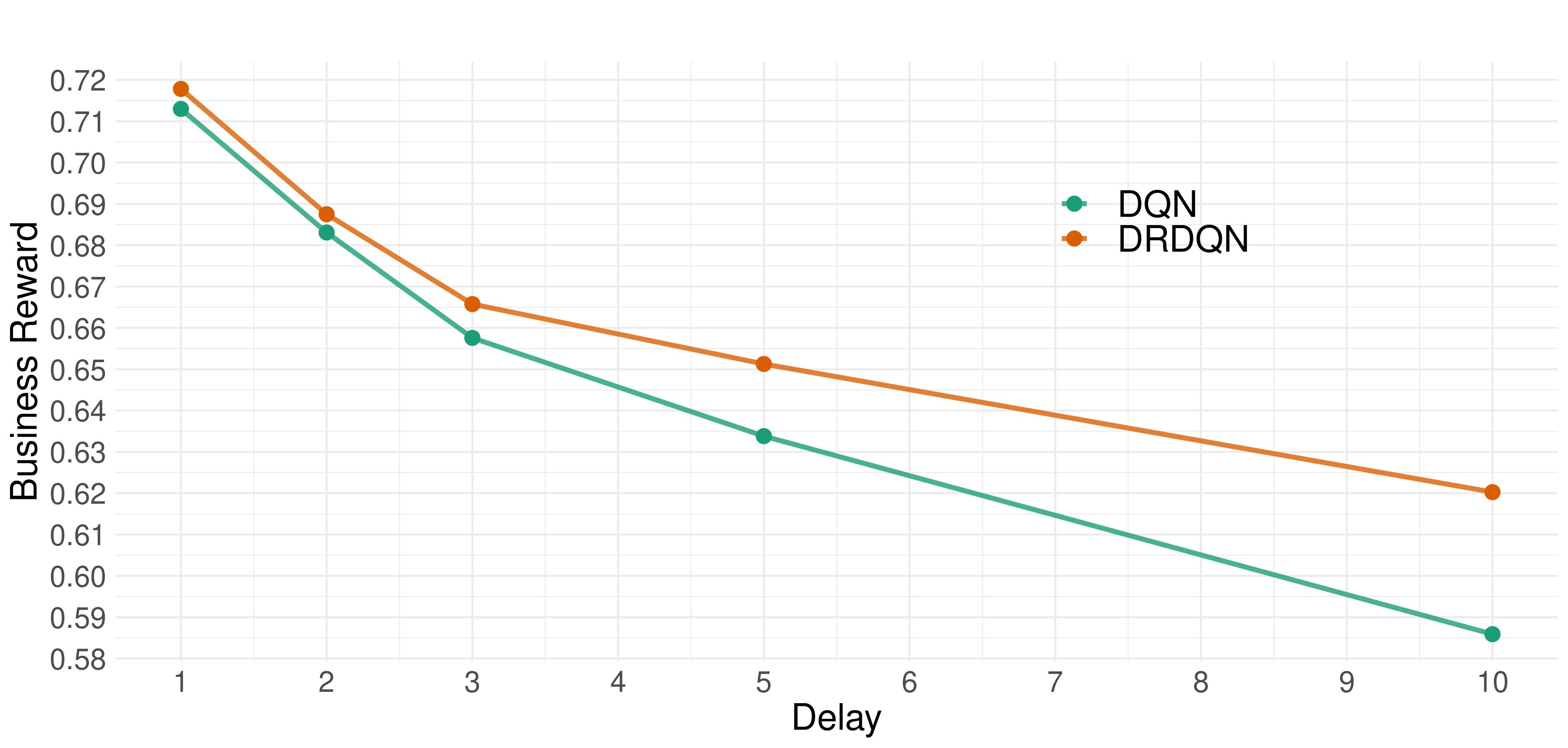}
\caption{Effect of lead time on 220 Product dataset}
\label{fig:220_effectDelay}
\end{figure}

\begin{figure}[H]
  \centering
  \includegraphics[width=0.99\linewidth]{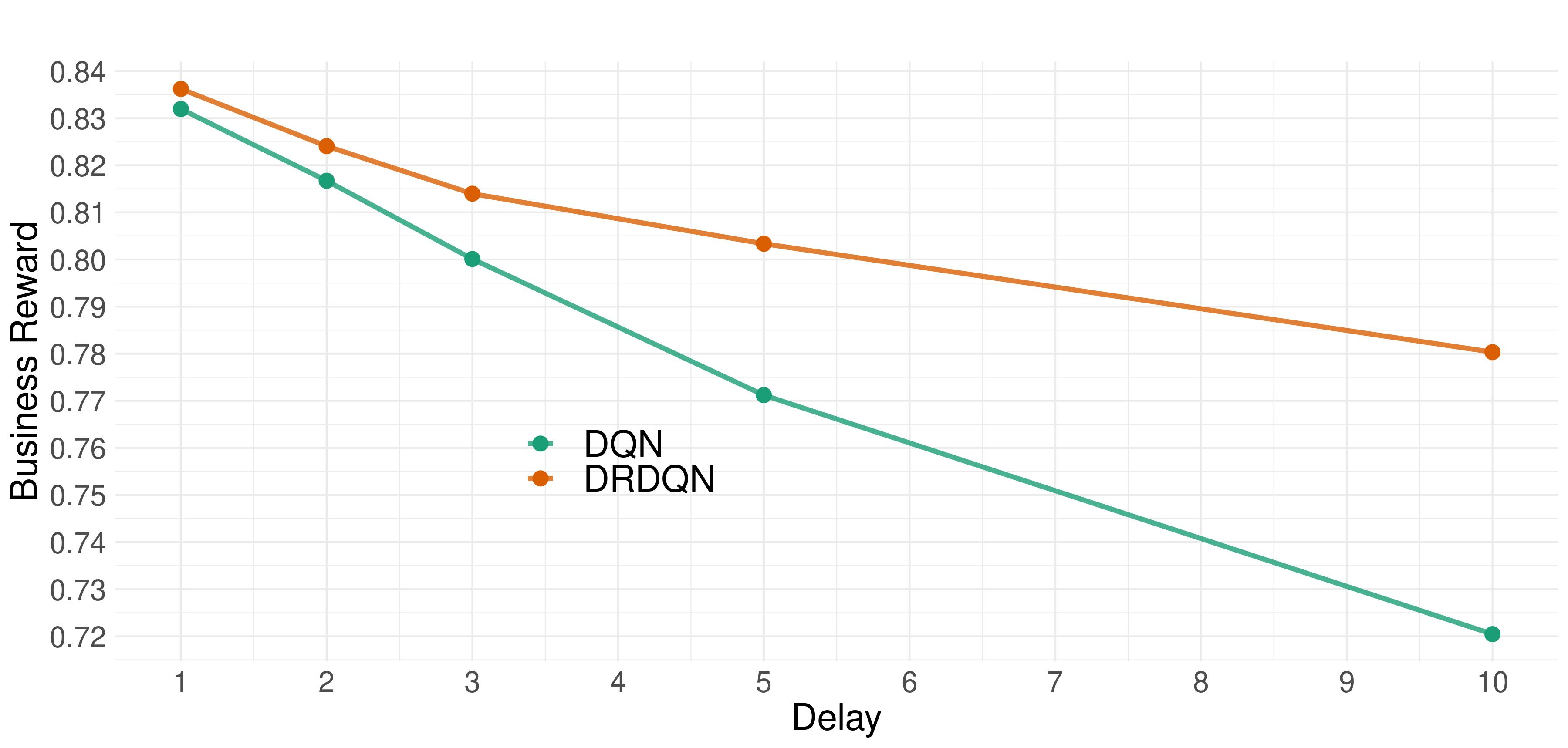}
\caption{Effect of lead time on 100 Product dataset}
\label{fig:100_effectDelay}
\end{figure}

Figures~\ref{fig:220_Training} and~\ref{fig:100_Training} shows the training graphs over 220 and 100 dataset respectively with different fixed lead times and
Figures~\ref{fig:220_effectDelay} and \ref{fig:100_effectDelay} show the effect of lead times between DQN and DRDQN. From all the plots, it is evident that there is significant drop in performance (business reward) as the lead time increases for DQN where as the decrease is much less for DRDQN. For example, DRDQN with 10 lead time is similar in performance to DQN with half the lead time. Thus, DRDQN is a more robust algorithm to changes in delay and this can also be observed for stochastic delays, as explained in the subsequent paragraphs. The difference, though quite small in terms of magnitude of the business rewards, can be quite significant when the overall profit margins are considered across all the products.

\begin{figure}[H]
  \centering
  \includegraphics[width=0.99\linewidth]{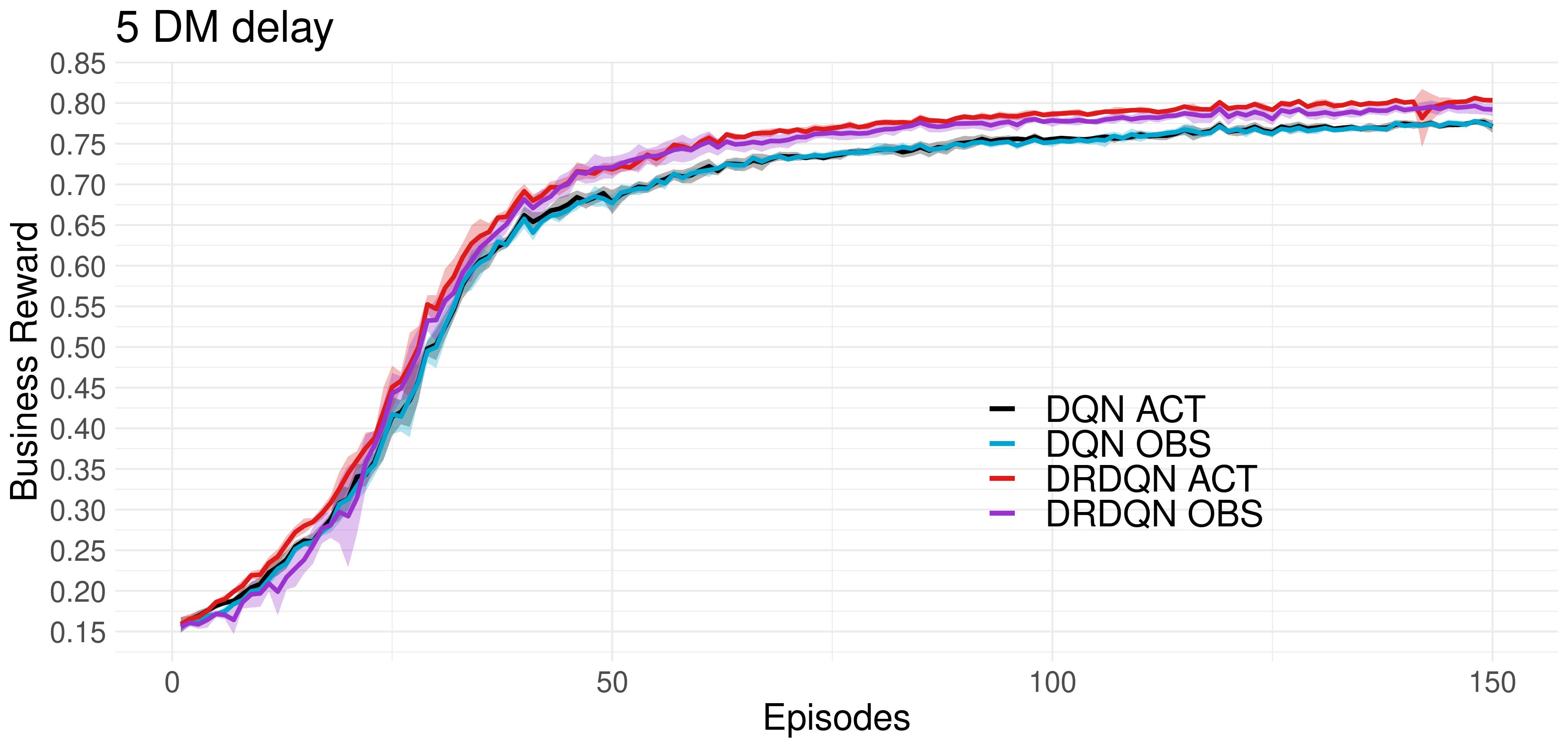}
\caption{Comparison of action and observation delay on 100 product dataset with 5 Lead-time.}
\label{fig:ACT_OBS_eq}
\end{figure}

We have considered lead times as action delays, however we can also consider having observation delays where we have older states while taking a decisions but there are no delays in implementation of actions. Figure~\ref{fig:ACT_OBS_eq} shows the equivalence of action delays and observation delays which is inline with results reported in \cite{nath2021revisiting}. We can clearly observe that the DRDQN with action and observation delay is able to outperform DQN. Results are only reported for 100 product dataset with 5 delay both for action and observation. We have observed similar results across dataset and delays.

\begin{figure}[H]
  \centering
  \includegraphics[width=0.99\linewidth]{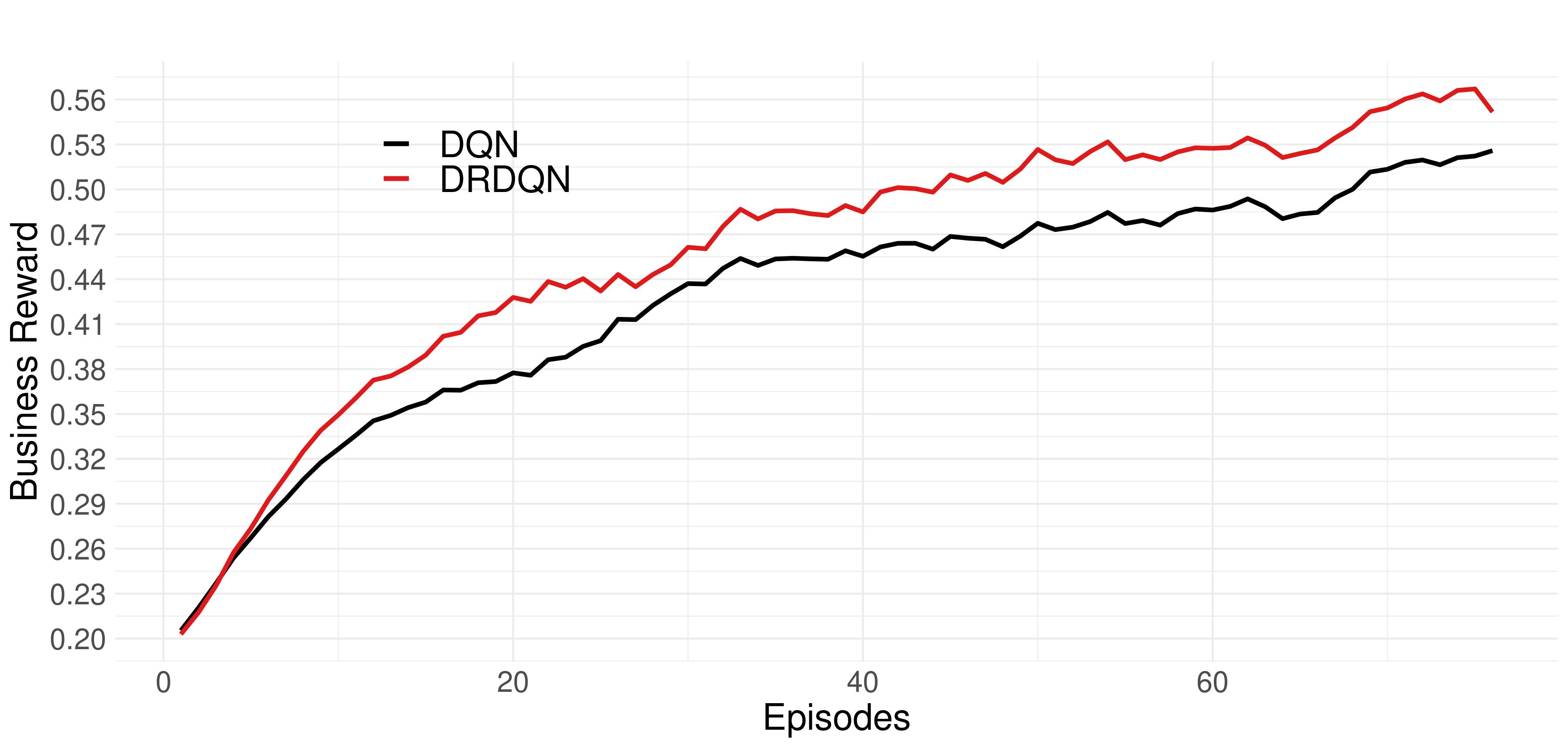}
\caption{Stochastic delay of maximum 50 lead time on 220 Product dataset, graph is plotted with moving average of 20}
\label{fig:220_Stochastic50}
\end{figure}

We have also experimented with stochastic lead time scenario. For this as mentioned earlier, stochasticity is across the episode whereas for a particular episode lead time remains constant. Figure~\ref{fig:220_Stochastic50} shows the training results for 220 product dataset where value of lead time for each episode was chosen at random between 1 and 50.

\section{Conclusion}
This paper addresses a key issue on managing uncertainty in lead times and real-time information sharing across multiple echelons in a supply chain system for optimizing inventory replenishment. The proposed work leverages the recently introduced delay-resolved framework in the RL literature to account for stochasticity at all levels in a computationally efficient manner, where lead times are viewed as action delays associated with the supply chain system. Unlike the existing RL-based inventory management system, our framework requires \emph{only} one agent to be trained for different values of lead times. In doing so, the model simply augments the past stocking decisions to its information state without needing to work with forecasted demands or policy roll-outs. Thus, the model can be trained efficiently and scaled suitably to account for cross-product constraints. To the best of authors' knowledge, this is the first such work that concurrently takes into account all aspects of a supply chain inventory control in a computationally efficient manner.

\bibliography{aaai22.bib}
\end{document}